\documentclass[a4paper]{spie}  

\usepackage{amsmath,amsfonts,amssymb}
\usepackage{graphicx}
\usepackage{subcaption}
\usepackage[colorlinks=true, allcolors=blue]{hyperref}
\usepackage{hyperref}

\usepackage{xcolor}
\usepackage{gensymb}
\usepackage{enumitem}
\usepackage{wrapfig}
\usepackage{graphicx}
\usepackage{comment}
\title{Segmenting Microcalcifications in Mammograms and its Applications}

\author[a]{Roee Zamir}
\author[b]{Shai Bagon}
\author[c]{David Samocha}
\author[c]{Yael Yagil}
\author[b]{Ronen Basri}
\author[c]{Miri Sklair-Levy}
\author[b]{Meirav Galun}
\affil[a]{\small Weizmann Institute of Science, Israel}
\affil[b]{\small Weizmann Artificial Intelligence Center (WAIC), Israel}
\affil[c]{\small Department of Diagnostic Imaging, Chaim Sheba Medical Center, Sackler School of Medicine, Tel-Aviv University}

\begin{document} 
\maketitle

\begin{abstract}
Microcalcifications are small deposits of calcium that appear in mammograms as bright white specks on the soft tissue background of the breast. Microcalcifications may be a unique indication for Ductal Carcinoma in Situ breast cancer, and therefore their accurate detection is crucial for diagnosis and screening. 
Manual detection of these tiny calcium residues in mammograms is both time-consuming and error-prone, even for expert radiologists, since these microcalcifications are small and can be easily missed. 
Existing computerized algorithms for detecting and segmenting microcalcifications tend to suffer from a high false-positive rate, hindering their widespread use. In this paper, we propose an accurate calcification segmentation method using deep learning. We specifically address the challenge of keeping the false positive rate low by suggesting a strategy for focusing the hard pixels in the training phase. Furthermore, our accurate segmentation enables extracting meaningful statistics on clusters of microcalcifications.

\end{abstract}



\section{Introduction}
\begin{figure}
  \centering
  
  \begin{tabular}{cc}
    \begin{minipage}{0.6\linewidth}\includegraphics[width=\linewidth]{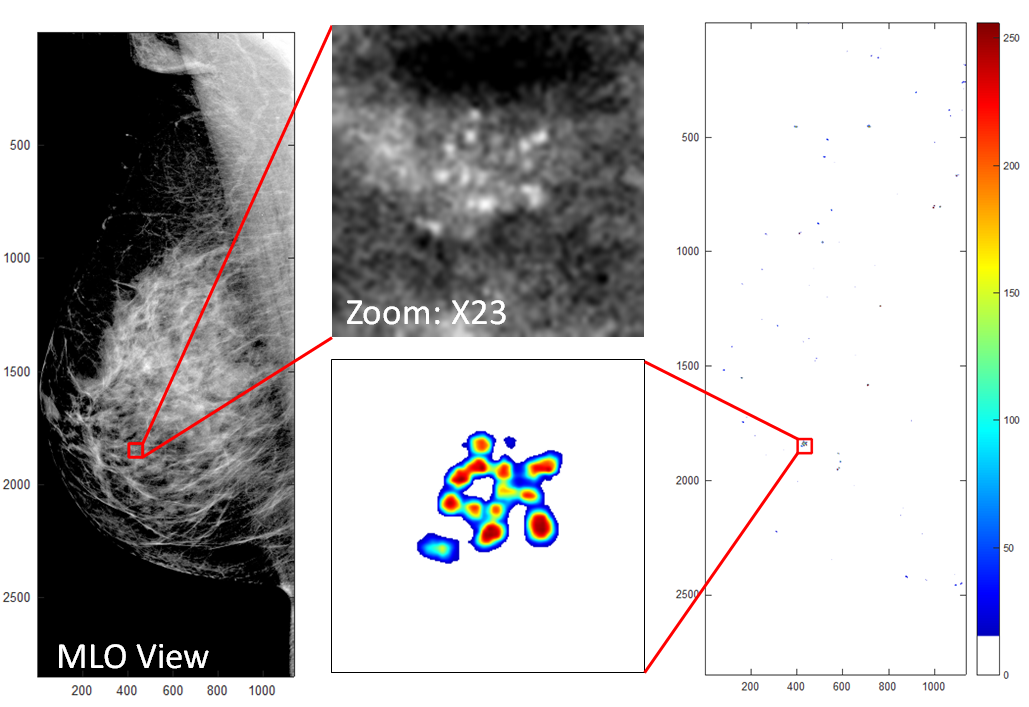}\end{minipage} &

    \begin{tabular}{|l|c|}
      \multicolumn{2}{c}{Shape Statistics} \\
      \hline \hline
      \# Calcifications & 13 \\
      \hline
      Mean area [mm$^2$] & 1.22\\  
      \hline
      Area STD [mm$^2$] & 0.89 \\
      \hline
    \end{tabular}
  \end{tabular}

    \caption{
    \emph{Left:} MLO Mammogram with a cluster of micro-calcifications marked with a red box.
    \emph{Middle:} Our segmentation results on the entire mammogram, and for the marked cluster.
    Our segmentation can save time, errors and eye fatigue in this otherwise tedious task. 
    \emph{Right:} Our accurate segmentation allows for automatically extracting meaningful shape statistics for clusters of microcalcifications, as shown in the table on the right.
    }
    \label{fig:galaxy} 
\end{figure}

Breast cancer is the most common cancer in women worldwide~\cite{ref1,ref2} and the second most common cause of cancer death~\cite{ref3}. Mammography is an X-Ray examination of the breast, and it is the main modality used for early detection and screening.
Since the 1970's when the mammography was introduced breast cancer mortality was substantially reduced~\cite{ref4,ref5}. 

One type of lesion observed in mammograms is calcifications.
Calcifications are small deposits of calcium that appear as bright specks or dots within the soft breast tissue, see Fig.~\ref{fig:galaxy}.
With some characteristics, microcalcifications can indicate an early stage of Ductal Carcinoma in Situ. 
In some cases these tiny specks are the only indication~\cite{ref12}, and thus must be detected. 
Mammogram reading done by radiologists. The reading process is time consuming, tedious and error-prone~\cite{ref10,ref6} since microcalcifications are often very small and easy to miss. This is in contrast to \emph{macro}calcifications, which are easy to spot and commonly indicate benign findings. Early detection of micro-calcification clusters (see, e.g., Fig~\ref{fig:galaxy}) is important for timely diagnosis of Ductal Carcinoma in Situ breast cancer.

In this paper we address the task of detecting and segmenting microcalcifications while maintaining low false-positive rate. 
Furthermore, accurate segmentation facilitates the extraction of important statistics such as the number of calcifications in a cluster, their size and variability (see, e.g., table in Fig~\ref{fig:galaxy}). The statistics, in turn, allow easier detection of changes in follow-up examinations.
Our approach is based on  a fully convolutional neural network (FCN).
Our network processes each mammogram at its original high-res scale, avoiding any sub-sampling so as not to miss small microcalcifications.
The challenge in this high-res processing is twofold: 
keeping the computation burden at bay, 
while at the same time keeping the false positive rate very low.
To illustrate the importance of achieving extremely low false positive rate, consider a mammogram at a typical 10 MPixel resolution:
even if we achieve a false positive rate of 0.01\% (i.e., one false alarm in 10,000) we would still get 1,000 falses \emph{per scan}!
To address these challenges we design a slim deep neural architecture that can process high-res mammograms efficiently and quickly.
Moreover, we employ strict online hard negative mining 
in training to ensure very low false alarm rate.

Our design focuses on the delineation of microcalcification, rather than providing a single, ``global" classification of the entire mammogram. Furthermore, accurate segmentation facilitates easier comparison between follow-up screenings of the same patient. To train our FCN model, we curated a new dataset of mammograms of different patients of Sheba Medical Center. 
This new dataset attributes special emphasis to microcalcifications: 
it contains many examples of both isolated as well as clusters of microcalcifications. 
Moreover, our dataset has \emph{pixel-wise} labeling of all microcalcifications.

To summarize, this work makes the following contributions:
\begin{itemize}[label=\small$\bullet$]
    \item A FCN for detecting \emph{micro}calcifications in breast mammograms with high accuracy and low false detection rate.
    \item Fine segmentation which allows extraction of important statistics (as number of calcifications in a cluster, their size etc.) and comparison between corresponding regions in pairs of mammograms of the same patient from different times.  
    \item A new dataset focusing on the challenge of microcalcification segmentation. 
\end{itemize}

\section{Related Work}
Detecting microcalcifications in mammograms is the subject of many studies.
Early approaches used morphological operations~\cite{classic1,classic2}, wavelets~\cite{classic3,classic4,classic5}, fractal models~\cite{classic7,classic8}, Laplacian scale-space representations~\cite{classic6}, and more recently machine learning schemes based on SVM with hand-crafted features~\cite{svm1,svm2}. 
These classic methods, however,  are prone to high false positive rates due to the small sizes of microcalcifications.

Recent work applied deep learning tools. 
Wang et~al.~\cite{nature1} used a semi-automated segmentation method to characterize all microcalcifications. 
Valvano et~al.~\cite{ref18} proposed a two-stage pipeline: a region proposal CNN first suggests candidate regions, then a classification CNN classifies each proposal. 
Hossain~\cite{unetmamm} proposed a method composed of multiple preprocessing stages followed by manual selection of suspecious regions, to be then fed to a trained segmentation U-Net~\cite{unet}. 
Our method, in contrast, is based on a simple lightweight network which segments microcalcifications in full size mammograms at \emph{a single} forward pass without the need of manual guidance.
Moreover, our method does not require any preprocessing e.g., cropping of suspicious patches or segmenting out the Pectoral muscle as some of the other methods require.

Developing a reliable computer aided diagnosis (CAD) system requires suitable labeled training data.
Only a few publicly available datasets of mammograms are currently available. 
The two most well-known are DDSM~\cite{ddsm} and MIAS~\cite{mias},
but they consist of screen-film mammography (SFM) images, which is a deprecated analog technology.
A more recent dataset, INbreast~\cite{inbreast} consists of full-field digital mammography (FFDM) images, some of which show microcalcifications.
Nevertheless, its annotations only provide accurate segmentation for \emph{isolated} microcalcifications, but not for \emph{clusters} of microcalsifications.
To address this lacuna we collaborated with Sheba Medical Center and curated a new dataset comprised of 42 FFDM mammograms of patients who have both isolated as well as clusters of microcalcifications.
Our data is carefully annotated with accurate segmentations of microcalcifications of both kinds.
Training deep networks using our new data results with better performance: improved accuracy without compromising on low false alarm rates.

\section{Method}
The task of pinpointing microcalcifications in full resolution mammogram can be as challenging as finding a needle in a haystack. 
In fact, microcalcifications can be as small as 0.5mm, covering an area of very few pixels in a mammogram of $\sim$10 MPixels (whose pixel resolution is about 100$\mu$m).
Fig.~\ref{fig:galaxy} exemplifies this challenge: The calcium residues are tiny and hard to distinguish from their surrounding tissues. 
Moreover, it seems that calcification residues are local phenomena; they have little effect on surrounding tissue, and they can be formed almost anywhere in the breast. 
This is in contrast to lesions and tumors that often affect the surrounding tissue and blood vessels and thus leave ``contextual marks" in their vicinity. 
This property makes localization of microcalcifications yet more challenging as there are no indications for the presence of a residue apart from the very tiny speck itself. 
Hence, any detection mechanism has to scrutinize the entire breast tissue at a very fine resolution.

\begin{figure}
    \centering
    \begin{tabular}{cc}
    \includegraphics[width=0.5\linewidth]{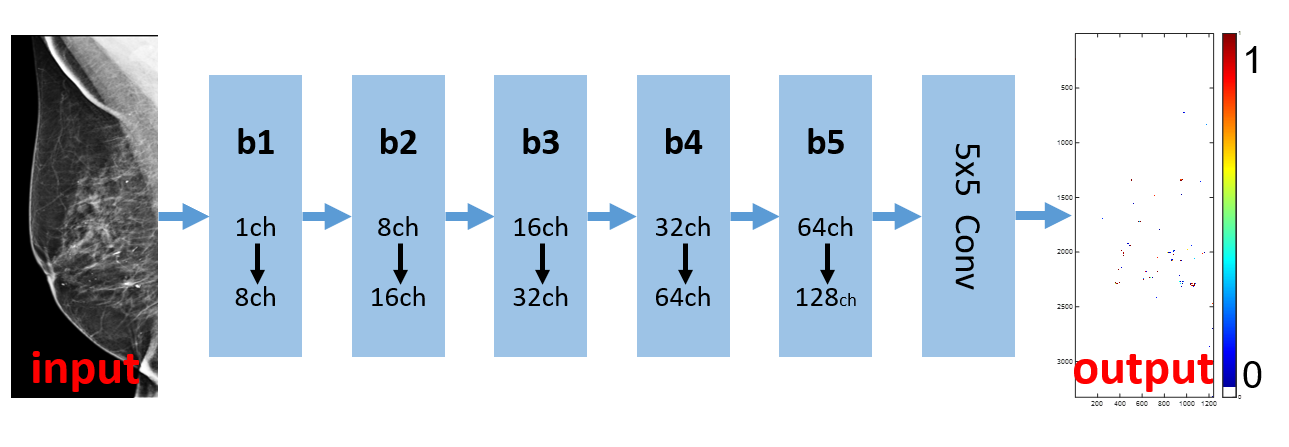} &
    \includegraphics[width=0.2\linewidth]{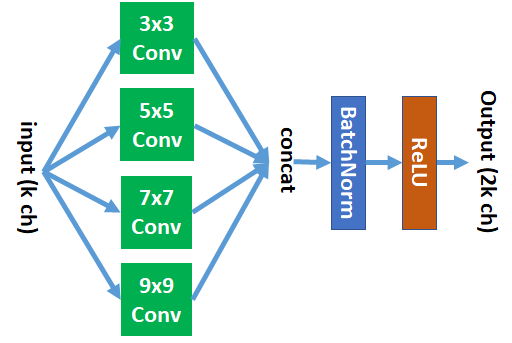} \\
    (a) Our FCN architecture. & (b) A basic block. 
    \end{tabular}
    \caption{(a)~Our model comprises 5 basic blocks connected to each other with an additional $5\times 5$ convolution on top.
    (b)~A detailed diagram of our basic block.
    \label{fig:netarch} }
\end{figure}

\subsection*{Model}
In contrast to popular deep models designed with large receptive fields and relying on contextual cues (e.g., \cite{deeplabv32017,fpn,unet}), we specifically design our FCN model to have a small receptive field.
The rationale behind this design choice is that calcification residues are local phenomena and therefore contextual information does not provide discriminative information.
Fig.~\ref{fig:netarch} shows the architecture of our FCN.
Inspired by Inception modules \cite{inception}, the basic building block of our model has 4 parallel 2D convolutions of varying kernel sizes.
This allows for capturing both \emph{micro} as well as \emph{macro} calcifications.
The output of all four convolutions is then concatenated to form the output of the block.
An additional $5\times 5$ conv layer outputs a scalar prediction score per-pixel.
This design is very ``slim": it has only $\sim$450K trainable parameters and a receptive field of $45 \times 45$ pixels allowing it to rapidly spot minute details in high-res mammograms.

\subsection*{Training}
The main difficulty in training a pixel-wise segmentation model of tiny microcalcifications in high-res mammograms is the severe class/non-class imbalance: there are about 1K negatives for each positive pixel.
As a result, the gradients of the binary cross-entropy loss function are overwhelmed by the majority negative class in a way that masks important but weak updates for the positive class.  
Na\"ively training a model under these conditions simply yields a constant ``negative" prediction, which is mostly accurate, but meaningless.
We use online hard negative mining \cite{hardnegativemining} as an integral part of our training procedure, computing the gradients of the loss function of only few hard negative examples in each iteration.
We maintain a ratio of 1:3 between positive and hard negative examples discarding the gradients from all the rest of the negatives.
This way gradients resulting from positive examples contribute roughly equally to negative examples allowing the model to converge to non-trivial accurate predictions.
Note that the imbalance in this case is so severe that simple modifications to the loss function (e.g., \cite{focalloss}) were not effective.

To further enrich our training examples we used random data augmentations: 
rotation within the interval $\left[-35\degree, 35\degree\right]$ and horizontal flipping.
We trained our model with Stochastic Gradient Descent (SGD) with momentum 0.9 and a fixed learning rate $\lambda=0.001$ for 500-1000 epochs.

\section{Experimental Results}
\begin{table}[t!]
\centering
\caption{Datasets of annotated full-field digital mammography (FFDM).\label{tab:datasets}}
\begin{tabular}{|l|c|c|c|c|}
\hline
DB Name & \#Patients & \#CC & \#MLO  & Resolutions (pixels)\\
\hline
INbreast & not-disclosed & 56 & 62 &  3328x2560, 4084x3328\\ 
Sheba  & 19 & 20 & 22 &  2294x1914, 2850x2394, 3062x2394, 4096x3328 \\
\hline
\end{tabular}
\end{table}
\begin{figure}
    \centering
        \begin{tabular}{cc}
        \includegraphics[width=0.33\linewidth]{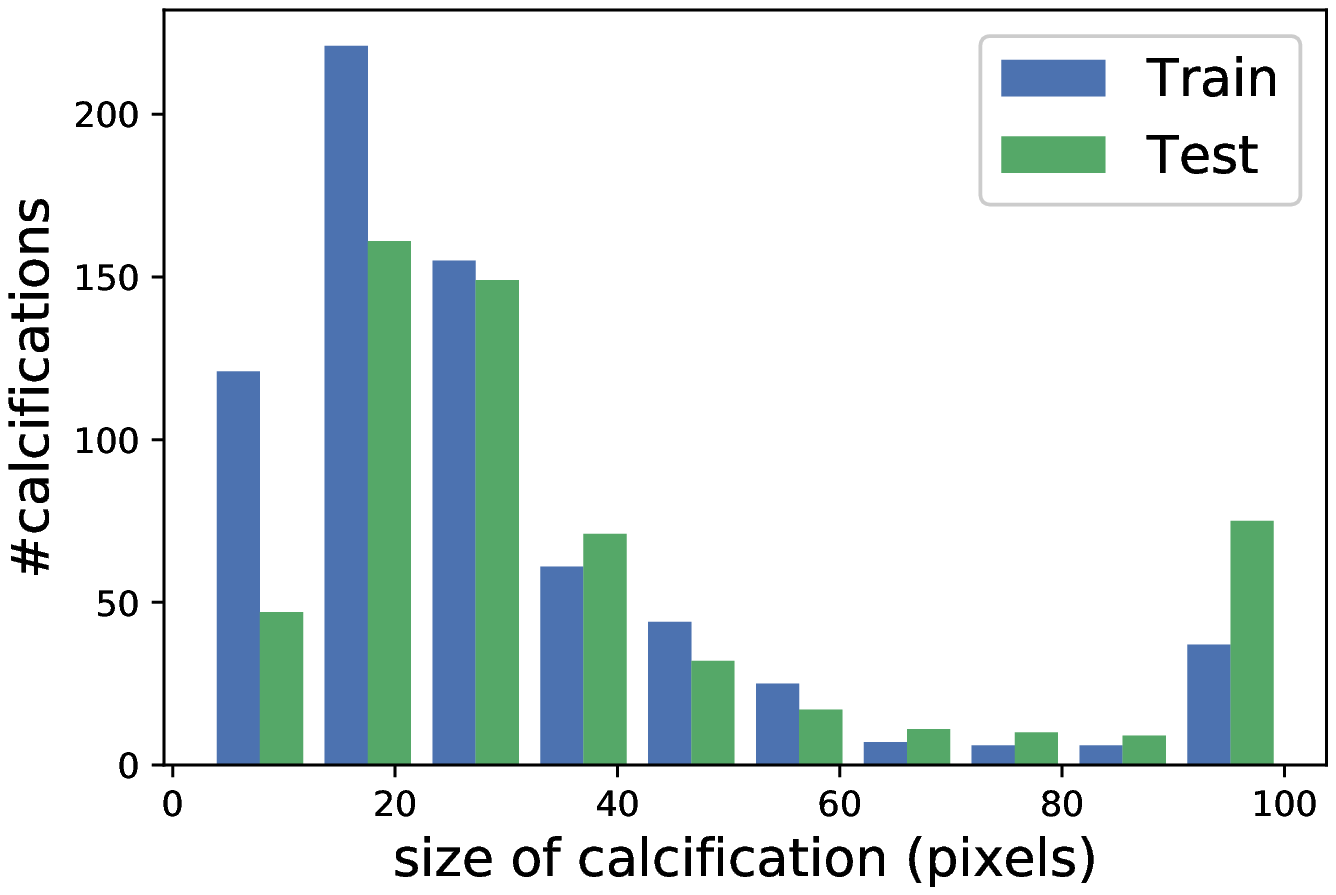} &
        \includegraphics[width=0.33\linewidth]{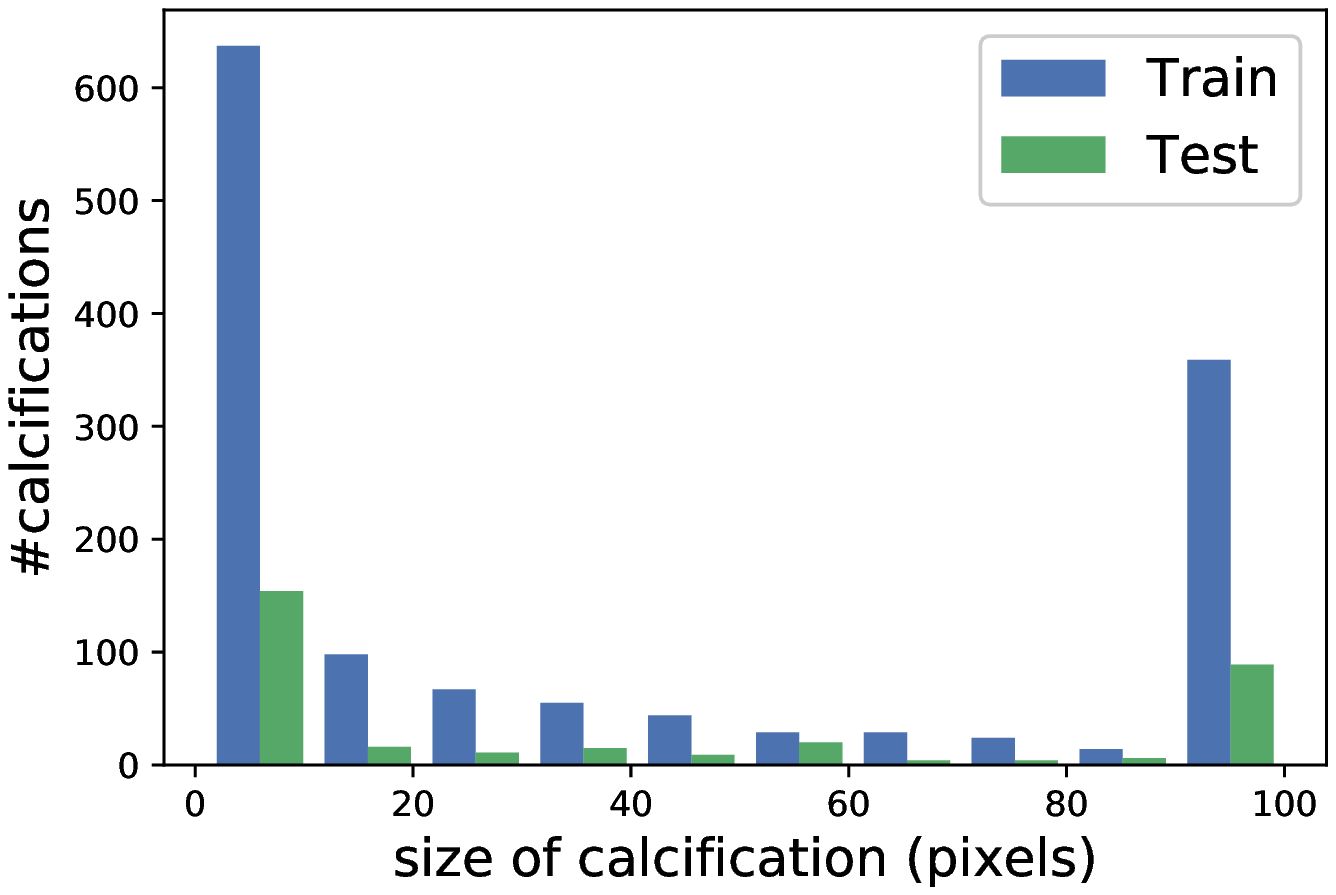} \\
        (a) Sheba dataset & (b) INbreast dataset
        \end{tabular}
        \caption{Truncated histogram of calcification size (rightmost bin is for size $\ge 100$).}
        \label{fig:both_hist}
\end{figure}

\noindent\textbf{Data}. \quad
A screening mammographic examination usually consists of two views per breast: craniocaudal (CC) which is top-down view and mediolateral oblique (MLO) which is a projection in a $45\degree$ angle.
Here we use both views to train and test our model -- as calcifications tend to appear the same under these different views.
We used two datasets for our experiments:

\noindent\textbf{INbreast Dataset \cite{inbreast}}
is a comprehensive annotated dataset of FFDM mammograms. It has accurate annotations of \emph{isolated} macro and microcalcifications. 

\noindent\textbf{Sheba Medical Center Dataset:}
Although very comprehensive, INbreast dataset lacks accurate annotations of \emph{clusters} of microcalcifications, which are very important bio markers for early stage Doctal Carcinoma in Situ.
To address this lacuna we curated a new dataset comprised of 42 mammograms of patients who have both isolated as well as clusters of microcalcifications.

Table~\ref{tab:datasets} provides a concise description of both INbreast and our new dataset. We use the pixel-wise annotations of the two datasets to train our model to output a ``calcification" probability per pixel in the input mammogram. We randomly split each dataset into 75\% train and 25\% test.
Fig.~\ref{fig:both_hist} shows the distribution of calcification sizes in both splits. The mammograms of the two datasets were acquired using different machines and were digitized using different pipelines, hence models trained on data from one dataset cannot be used on the other and vice versa. 
Therefore, we use the same architecture and training process but train a unique model-parameters for each dataset independently. 



\subsection{Segmentation}

\begin{figure}[b!]
\centering
\includegraphics[width=\linewidth]{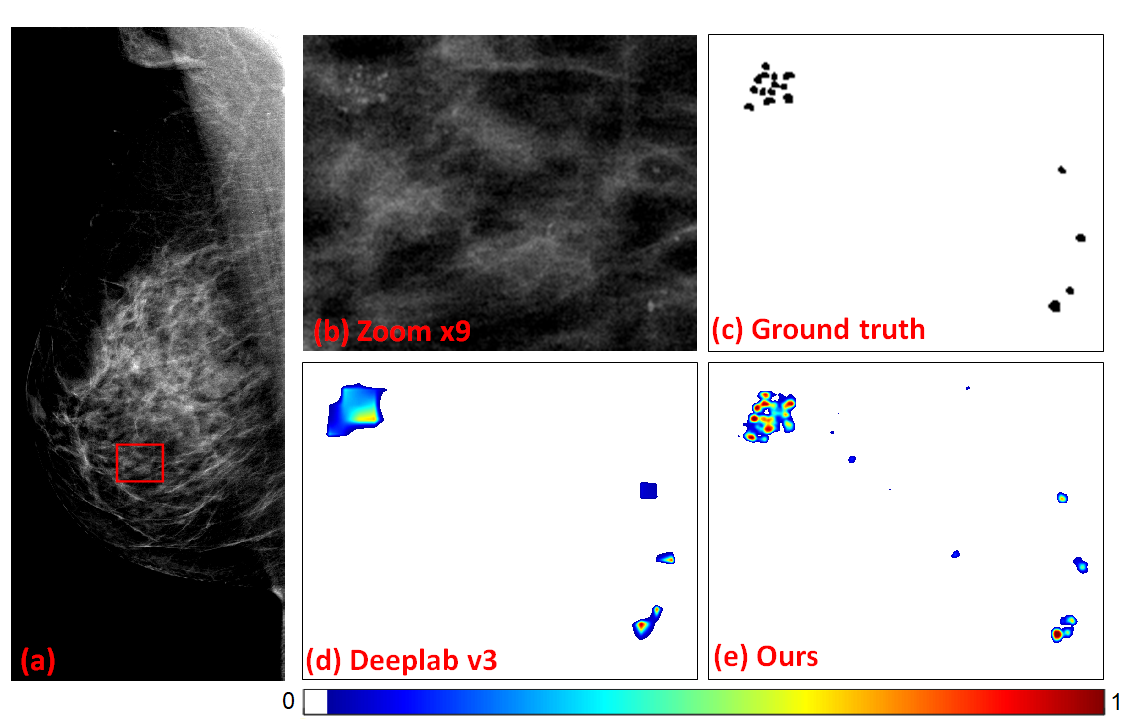}
\caption{Qualitative comparison: (a) Input MLO mammogram from Sheba dataset. 
(b) $\times 9$ zoom on marked red region (c) Ground truth (d) Baseline DeepLabV3 result (e) Ours.
Our model can accurately delineate both isolated microcalcifications as well as densely clustered ones.} \label{fig:comparison}
\end{figure}  
\begin{figure}[!t]
\centering
\includegraphics[width=1\linewidth]{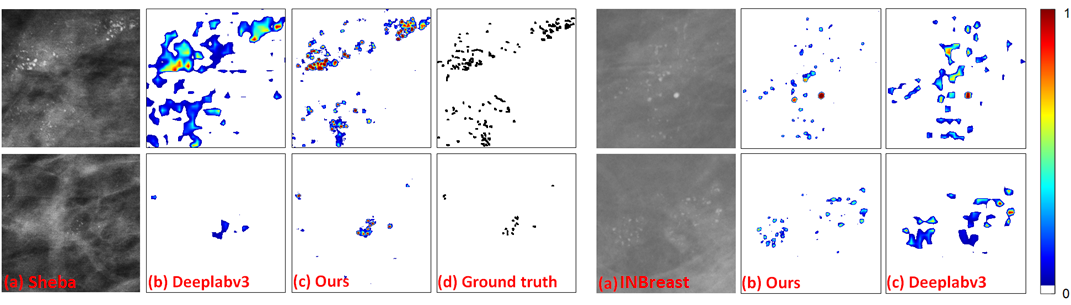}
\caption{Our results vs DeepLabV3 results on two Regions Of Interest from the Sheba (left) and INBreast (right) datasets. Note that our output is more accurate and facilitates the extraction of meaningful statistics as shown in Fig~\ref{fig:galaxy} and~\ref{fig:temporal_comparison2}.\label{fig:inbreast_pred_vis}}
\end{figure}  

\begin{figure}
    \centering
    \includegraphics[width=\linewidth]{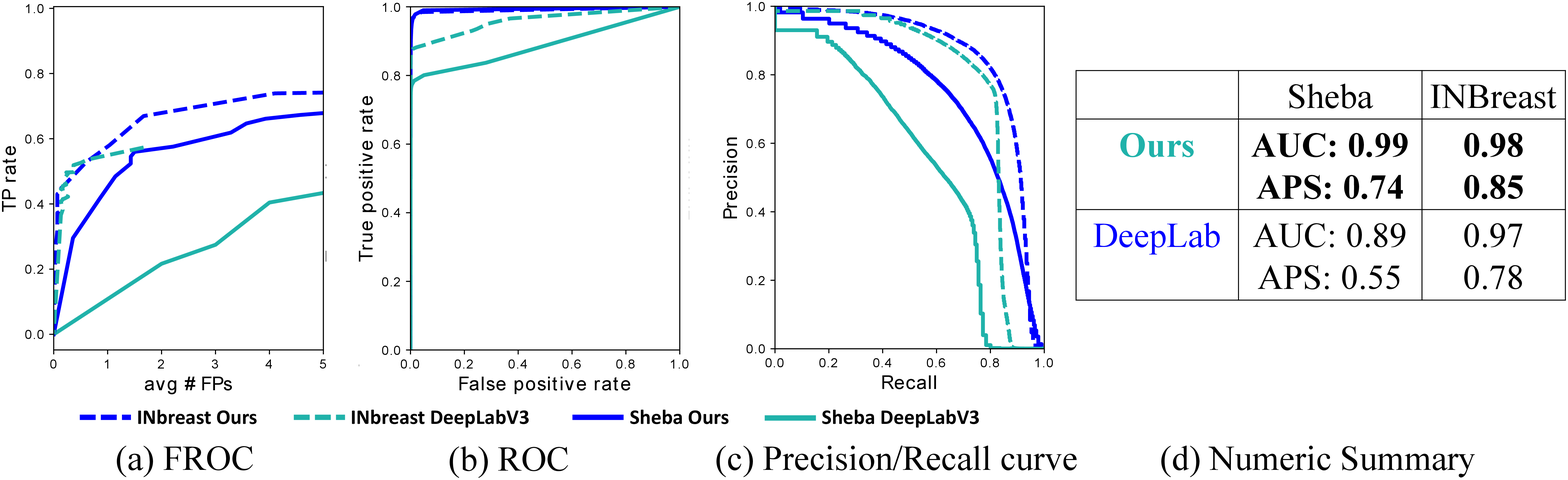}
    \vspace*{.2cm}
    \caption{Comparison on our method to DeepLabV3 baseline on both INbreat data and ours. (a)~FROC curve for ``microcalcification level" evaluation. (b)~ROC curve for pixel-level evaluation. (c)~Precision - Recall graph.
    (d)~Area Under Curve (AUC) and Average Precision (APS).}
    \label{fig:results_curves}
\end{figure}

As a baseline for comparison we fine-tune state-of-the-art DeepLabV3 \cite{deeplabv32017} semantic segmentation model.
The baseline DeepLabV3 model has $\sim$40\textbf{M} trainable parameters (two orders of magnitude more than ours) and a receptive field of roughly $900\times 900$ pixels ($\sim\times 200$ than ours).
We were unable to compare to \cite{nature1,ref18}, since they provide neither code nor data for comparison.

Fig.~\ref{fig:comparison} and ~\ref{fig:inbreast_pred_vis} show segmentation examples.
Our model outputs per-pixel ``calcification probability" in the range of $\left[0, 1\right]$ depicted as a heat map from blue (low probability) to red (high probability) in the figures.
Our method delineates the microcalcifications more accurately than the baseline.
More importantly, in clusters of microcalcifications DeepLab's results are ``smeared" and do not allow to distinguish between the different microcalcifications. 
In our segmentation, on the other hand, one can distinguish between the different microcalcifications, which is an important feature for diagnosis.


We further used the annotations of the test sets 
for quantitative evaluation.
First, we checked accuracy on a ``calcification level"; that is, we partitioned the annotations and the predictions according to their connected components. 
We treat each such component as a single calcification. 
Fig.~\ref{fig:results_curves}~(a) shows FROC \cite{froc} measuring the calcification detection rate as a function of the average number of false alarms per mammogram. 
Our model performs better than the baseline detecting more microcalcifications for any given false alarm rate.

Measuring accuracy at the calcification level is good for assessing performance on isolated microcalcifications where pixel-level accuracy is not required as long as all lesions are being detected.
However, when dealing with dense clusters of microcalcifications accurate \emph{pixel-level} prediction is crucial, see, e.g. Fig.~\ref{fig:comparison}.
Therefore, we further computed ROC and precision-recall curves measuring \emph{per-pixel} accuracy as shown in Fig.~\ref{fig:results_curves}~(b)~and~(c).
Due to the severe imbalance between the classes, the precision-recall curve is more meaningful as it is not affected by the size of the negative class. The table in Fig~\ref{fig:results_curves}.d
summarizes the graphs of Fig.~\ref{fig:results_curves}~a-c by showing the Area Under Curve (AUC) of the ROC curve and Average Precision Score (APS) of the precision-recall curves \cite{aps} (in both, higher is better). As can be seen, our method outperforms the baseline on both datasets in all performance measures.

\subsection{Tracking Changes over Time}
\begin{figure}[!t]
  \centering
  \begin{tabular}{cc}
    \begin{minipage}{0.65\linewidth}\includegraphics[width=\linewidth]{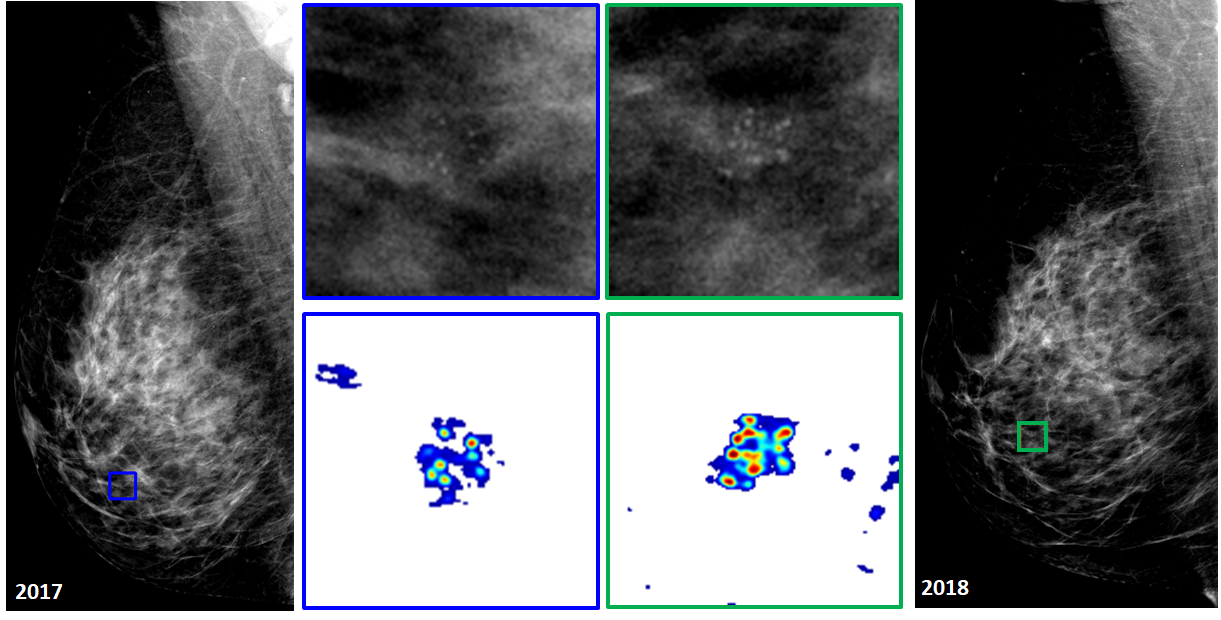}\end{minipage} &  
\begin{tabular}{|l|c|c|}
\multicolumn{2}{c}{Shape Statistics} \\
\hline \hline
& 2017 & 2018 \\
\hline
\#Calcifications & 7 & 11\\
\hline
Mean area [mm$^2$] & 0.57 & 0.59\\  
\hline
Area STD [mm$^2$] & 0.23 & 0.61\\
\hline
\end{tabular}
\end{tabular}
    \caption{Accurate segmentation allows to easily compare between mammograms of the same patient from different times (2017 and a follow-up on 2018). 
    Notice how our accurate segmentation enhance the temporal difference, as exemplified by the automatically extracted shape statistics.}
    \label{fig:temporal_comparison2}
    
\end{figure}

One important aspect of periodic screening is the ability to reliably and efficiently compare two scans of the same patient taken at different times.
Here we exemplify how our predicted segmentation can significantly assist clinicians in this otherwise tedious task.
Given two screening mammograms of the same patient taken in two different times,
the clinician marks a crude region around the suspicious cluster in both mammograms.
Our accurate segmentation facilitates easy comparison and progression assessment between the two scans.
As can be seen in Fig ~\ref{fig:inbreast_pred_vis}, our method segmenting the calcifications accurately while the baseline DeepLabV3 results are ``smeared" and do not allow to distinguish between the different microcalcifications.
Our accurate segmentation allows to \emph{automatically} extract important statistics (e.g., number of calcifications in a cluster, mean size etc.) to assess temporal changes in follow-up screenings as shown in Fig.~\ref{fig:temporal_comparison2}.

\section{Conclusion}
Calcification detection in mammograms is a tedious and time-consuming task that is currently done manually by the radiologists. 
In this work we proposed an efficient deep learning model trained to segment microcalcification accurately while maintaining low false positive rate.
Segmentation results on both the INBreast public dataset and our newly curated Sheba dataset show our methods exceeds DeepLabV3 baseline. 
Moreover our accurate segmentation allows for the automatic extraction of shape features from clusters of microcalcifications, assisting in change detection in follow-up examinations.


\subsection*{Acknowledgments}
Dr Bagon is a Robin Chemers Neustein Artificial Intelligence Fellow.
Additionally, Dr Bagon is supported by a research grant from the Carolito Stiftung. 

\bibliography{report} 

\begin{thebibliography}{10}

\bibitem{ref1}
``Global cancer statistics for the most common cancers,'' (2018).

\bibitem{ref2}
``Latest global cancer data,'' (2018).

\bibitem{ref3}
Bray, F. e.~a., ``Global cancer statistics 2018: Globocan estimates of
  incidence and mortali worldwide for 36 cancers in 185 countries,'' in [{\em
  CA Cancer J. Clin. 68}{\nolinebreak\hspace{0.1em}]},  (2018).

\bibitem{ref4}
Oeffinger, K. C. e.~a., ``Breast cancer screening for women at average risk:
  2015 guideline update from the american cancer society,'' in [{\em Guideline
  Update From the American Cancer Society. JAMA}{\nolinebreak\hspace{0.1em}]},
  (2015).

\bibitem{ref5}
Tabar, L., Yen, M.-F., Vitak, B., Hsiu-Hsi, Chen, T., Smith, R.~A., and Duffy,
  S.~W., ``Mammography service screening and mortality in breast cancer
  patients: 20-year follow-up before and after introduction of screening,'' in
  [{\em The Lancent}{\nolinebreak\hspace{0.1em}]},  (2003).

\bibitem{ref12}
Morgan, M., Cooke, M., and McCarthy, G., ``Microcalcifications associated with
  breast cancer: An epiphenomenon or biologically significant feature of
  selected tumors?,'' {\em J Mammary Gland Biol Neoplasia.}  (2005).

\bibitem{ref10}
CD, L., RF, A., BL, S., JM, L., DS, B., K, K., LM, H., T, O., AN, T., GH, R.,
  and DL, M., ``National performance benchmarks for modern screening digital
  mammography: Update from the breast cancer surveillance consortium,'' {\em
  Radiology.}  (2017).

\bibitem{ref6}
Barlow, W.~E., Chi, C., Carney, P.~A., Taplin, S.~H., D’Orsi, C., Cutter, G.,
  Hendrick, R.~E., and Elmore, J.~G., ``Accuracy of screening mammography
  interpretation by characteristics of radiologists,'' {\em Journal of the
  National Cancer Institute}~{\bf 96}(24),  1840--1850, Oxford University Press
  (2004).

\bibitem{classic1}
{Zhao}, D., {Shridhar}, M., and {Daut}, D.~G., ``Morphology on detection of
  calcifications in mammograms,'' in [{\em Proceedings of IEEE International
  Conference on Acoustics, Speech, and Signal
  Processing}{\nolinebreak\hspace{0.1em}]},  (1992).

\bibitem{classic2}
{Dengler}, J., {Behrens}, S., and {Desaga}, J.~F., ``Segmentation of
  microcalcifications in mammograms,'' {\em IEEE Transactions on Medical
  Imaging}  (1993).

\bibitem{classic3}
{Strickland}, R.~N. and {Hee Il Hahn}, ``Wavelet transforms for detecting
  microcalcifications in mammograms,'' {\em IEEE Transactions on Medical
  Imaging}  (1996).

\bibitem{classic4}
{Wang}, T.~C. and {Karayiannis}, N.~B., ``Detection of microcalcifications in
  digital mammograms using wavelets,'' {\em IEEE Transactions on Medical
  Imaging}  (1998).

\bibitem{classic5}
Qian, W., Clarke, L.~P., Kallergi, M., Li, H.~D., Velthuizen, R.~P., M.D., R.
  A.~C., and Silbiger, M.~L., ``{Tree-structured nonlinear filter and wavelet
  transform for microcalcification segmentation in mammography},'' in [{\em
  Biomedical Image Processing and Biomedical
  Visualization}{\nolinebreak\hspace{0.1em}]},  (1993).

\bibitem{classic7}
{Li}, H., {Liu}, K. J.~R., and {Lo}, S. .~B., ``Fractal modeling and
  segmentation for the enhancement of microcalcifications in digital
  mammograms,'' {\em IEEE Transactions on Medical Imaging}  (1997).

\bibitem{classic8}
Lefebvre, F., Benali, H., Gilles, R., Kahn, E., and Di~Paola, R., ``A fractal
  approach to the segmentation of microcalcifications in digital mammograms,''
  {\em Medical Physics}  (1995).

\bibitem{classic6}
{Netsch}, T. and {Peitgen}, H.~., ``Scale-space signatures for the detection of
  clustered microcalcifications in digital mammograms,'' {\em IEEE Transactions
  on Medical Imaging}  (1999).

\bibitem{svm1}
{El-Naqa}, I., {Yongyi Yang}, {Wernick}, M.~N., {Galatsanos}, N.~P., and
  {Nishikawa}, R.~M., ``A support vector machine approach for detection of
  microcalcifications,'' {\em IEEE Transactions on Medical Imaging}  (2002).

\bibitem{svm2}
{D'Elia}, C., {Marrocco}, C., {Molinara}, M., {Poggi}, G., {Scarpa}, G., and
  {Tortorella}, F., ``Detection of microcalcifications clusters in mammograms
  through ts-mrf segmentation and svm-based classification,'' in [{\em
  Proceedings of the 17th International Conference on Pattern Recognition,
  2004. ICPR 2004.}{\nolinebreak\hspace{0.1em}]},  (2004).

\bibitem{nature1}
Wang, J., Yang, X., Cai, H., Tan, W., Jin, C., and Li, L., ``Discrimination of
  breast cancer with microcalcifications on mammography by deep learning,''
  {\em Scientific reports}~{\bf 6}(1),  1--9 (2016).

\bibitem{ref18}
Valvano, G., Santini, G., Martini, N., Ripoli, A., Iacconi, C., Chiappino, D.,
  and Latta, D.~D., ``Convolutional neural networks for the segmentation of
  microcalcification in mammography imaging,'' {\em Journal of Healthcare
  Engineering}  (2019).

\bibitem{unetmamm}
Hossain, M.~S., ``Microcalcification segmentation using modified {U-net}
  segmentation network from mammogram images,'' {\em Journal of King Saud
  University - Computer and Information Sciences}  (2019).

\bibitem{unet}
Ronneberger, O., Fischer, P., and Brox, T., ``U-net: Convolutional networks for
  biomedical image segmentation,'' (2015).

\bibitem{ddsm}
Heath, M., Bowyer, K., Kopans, D., Moore, R., and Kegelmeyer, P., ``The digital
  database for screening mammography,'' {\em Proceedings of the Fourth
  International Workshop on Digital Mammography}  (2000).

\bibitem{mias}
Suckling, J., ``The mammographic image analysis society digital mammogram
  database''exerpta medica,'' (1994).

\bibitem{inbreast}
Moreira, I.~C., Amaral, I., Domingues, I., Cardoso, A., Cardoso, M.~J., and
  Cardoso, J.~S., ``{INbreast}: Toward a full-field digital mammographic
  database,'' {\em Academic Radiology}  (2012).

\bibitem{deeplabv32017}
Chen, L.-C., Papandreou, G., Schroff, F., and Adam, H., ``Rethinking atrous
  convolution for semantic image segmentation,'' {\em arXiv preprint
  arXiv:1706.05587}  (2017).

\bibitem{fpn}
Lin, T.-Y., Doll{\'a}r, P., Girshick, R., He, K., Hariharan, B., and Belongie,
  S., ``Feature pyramid networks for object detection,'' in [{\em Proceedings
  of the IEEE conference on computer vision and pattern
  recognition}{\nolinebreak\hspace{0.1em}]},  (2017).

\bibitem{inception}
Szegedy, C., Liu, W., Jia, Y., Sermanet, P., Reed, S., Anguelov, D., Erhan, D.,
  Vanhoucke, V., and Rabinovich, A., ``Going deeper with convolutions,'' in
  [{\em Proceedings of the IEEE conference on computer vision and pattern
  recognition}{\nolinebreak\hspace{0.1em}]},  (2015).

\bibitem{hardnegativemining}
Shrivastava, A., Gupta, A., and Girshick, R., ``Training region-based object
  detectors with online hard example mining,'' in [{\em Proceedings of the IEEE
  conference on computer vision and pattern
  recognition}{\nolinebreak\hspace{0.1em}]},  (2016).

\bibitem{focalloss}
Lin, T.-Y., Goyal, P., Girshick, R., He, K., and Doll{\'a}r, P., ``Focal loss
  for dense object detection,'' in [{\em Proceedings of the IEEE international
  conference on computer vision}{\nolinebreak\hspace{0.1em}]},  (2017).

\bibitem{froc}
Egan, J.~P., Greenberg, G.~Z., and Schulman, A.~I., ``Operating
  characteristics, signal detectability, and the method of free response,''
  {\em The Journal of the Acoustical Society of America}  (1961).

\bibitem{aps}
Manning, C.~D., Raghavan, P., and Sch{\"u}tze, H.,  [{\em Introduction to
  information retrieval}{\nolinebreak\hspace{0.1em}]}, Cambridge university
  press (2008).

\end{thebibliography}
\bibliographystyle{spiebib} 

\end{document}